\pdfoutput=1

\documentclass[11pt]{article}

\usepackage[final]{acl}

\usepackage{times}
\usepackage{latexsym}
\usepackage{xcolor}
\usepackage{tcolorbox}

\usepackage{tabularx}
\usepackage{booktabs}
\usepackage{enumitem}
\usepackage[T1]{fontenc}

\usepackage[utf8]{inputenc}

\usepackage{microtype}
\usepackage{amsmath}
\usepackage{amssymb}
\usepackage{colortbl}
\usepackage{booktabs}
\usepackage{pifont}
\usepackage{xcolor}
\usepackage{multirow}
\usepackage{inconsolata}

\usepackage{graphicx}
\definecolor{darkgreen}{rgb}{0.0, 0.5, 0.0}
\newcommand{\greencheck}{\textcolor{darkgreen}{\ding{51}}}

\hypersetup{
    colorlinks=true,
    linkcolor=red,
    citecolor=cyan,
    filecolor=magenta,      
    urlcolor=cyan,
    }
\title{
MedCoT: Medical Chain of Thought via Hierarchical Expert 
}

\author{
 \textbf{Jiaxiang Liu\textsuperscript{1}}\ \ \
 \textbf{Yuan Wang\textsuperscript{1}}\ \ \
 \textbf{Jiawei Du\textsuperscript{2, 3}}\ \ \
 \textbf{Joey Tianyi Zhou\textsuperscript{2, 3}}\ \ \
 \textbf{Zuozhu Liu\textsuperscript{*, 1}}
\\
\\
 \textsuperscript{1} \small ZJU-Angelalign R\&D Center for Intelligence Healthcare, Zhejiang University, China\\
 \textsuperscript{2} \small Centre for Frontier AI Research (CFAR), Agency for Science, Technology and Research (A*STAR), Singapore
\\
 \textsuperscript{3} \small Institute of High Performance Computing (IHPC), Agency for Science, Technology and Research (A*STAR), Singapore
\\
 { \small
   {\tt \{jiaxiang.21, zuozhuliu\}@intl.zju.edu.cn}
 }
}

\begin{document}

\maketitle

\begin{abstract}
Artificial intelligence has advanced in Medical Visual Question Answering (Med-VQA), but prevalent research tends to focus on the accuracy of the answers, often overlooking the reasoning paths and interpretability, which are crucial in clinical settings. Besides, current Med-VQA algorithms, typically reliant on singular models, lack the robustness needed for real-world medical diagnostics which usually require collaborative expert evaluation.
To address these shortcomings, this paper presents MedCoT, a novel hierarchical expert verification reasoning chain method designed to enhance interpretability and accuracy in biomedical imaging inquiries. MedCoT is predicated on two principles: \textit{The necessity for explicit reasoning paths in Med-VQA} and \textit{the requirement for multi-expert review to formulate accurate conclusions}. The methodology involves an Initial Specialist proposing diagnostic rationales, followed by a Follow-up Specialist who validates these rationales, and finally, a consensus is reached through a vote among a sparse Mixture of Experts within the locally deployed Diagnostic Specialist, which then provides the definitive diagnosis.
Experimental evaluations on four standard Med-VQA datasets demonstrate that MedCoT surpasses existing state-of-the-art approaches, providing significant improvements in performance and interpretability. 
Code is released at \url{https://github.com/JXLiu-AI/MedCoT}.
\let\thefootnote\relax\footnotetext{* Corresponding author.}
\end{abstract}

\section{Introduction}

\begin{figure}[t!]
\centering
\includegraphics[width=0.45\textwidth]{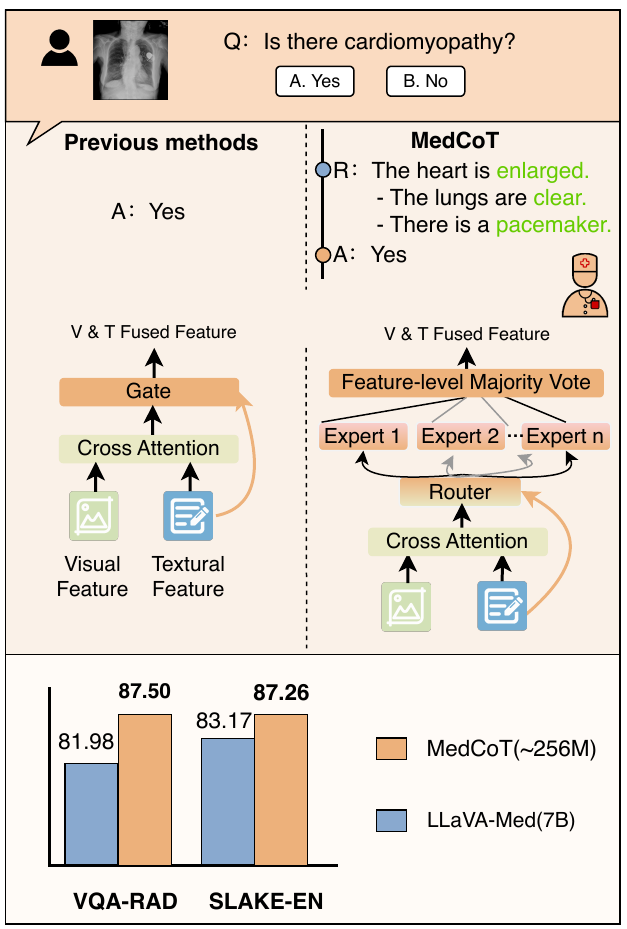}
\caption{
The upper figure shows a comparison of the outputs from the previous Med-VQA method and MedCoT, as well as the previous techniques in MMCoT \cite{zhang2023multimodal} versus Sparse MoE in MedCoT.
The lower figure demonstrates that MedCoT, with a model size of ~256M parameters, outperforms the 7B parameter LLaVA-Med by 5.52\% and 4.09\% (Accuracy) on the VQA-RAD and SLAKE-EN datasets. 
} 
\label{fig1}
\end{figure}

Medical Visual Question Answering (Med-VQA) has recently gained significant attention \cite{chen2022align, gong2021cross, ren2020cgmvqa, khare2021mmbert}. As a new exploration in the medical domain, Med-VQA aims to answer medical questions in natural language based on input medical images. An effective Med-VQA system can assist clinicians in interpreting medical images, thereby ensuring and accelerating the diagnostic process. For patients, automated Med-VQA services can greatly satisfy the demand for personalized health consultations \cite{liu2023parameter}.

In the field of Med-VQA, numerous attempts have been made using deep learning technologies \cite{tiong2022plug, banerjee2020weaqa, changpinyo2022all, liu2023chatgpt, gai2024medthink}.
For instance, \citet{nguyen2019overcoming} utilized Bilinear Attention Networks (BAN) \cite{kim2018bilinear} and enhanced them for Med-VQA by incorporating a Mixed Enhanced Visual Feature (MEVF) setup consisting of pre-trained meta-learning modules and Convolutional Denoising Autoencoders (CDAE). Building on this, \citet{zhan2020medical} designed a conditional reasoning framework to boost the inference capabilities of Med-VQA models. However, these approaches often underperform in many practical scenarios, primarily due to poor capabilities in extracting and integrating features from a limited number of medical images and text data \cite{eslami2021does, song2022clip, wang2022clip}.
\citet{eslami2021does} introduced the CLIP architecture into the framework by deploying it as the visual encoder within MEVF \cite{nguyen2019overcoming}, pre-trained on the multimodal medical dataset ROCO \cite{pelka2018radiology}. Their experiments demonstrated significant improvements with the CLIP. \citet{liu2023parameter} developed VQA-Adapter, which uses a lightweight adapter and label smoothing to efficiently fine-tune the CLIP model for Med-VQA, thus reducing computational costs and mitigating overfitting.
\citet{li2024llava} proposed LLaVA-Med, which utilizes GPT-4 and a novel curriculum learning approach to efficiently train LLaVA on biomedical images, significantly enhancing Med-VQA capabilities.

However, previous Med-VQA approaches typically focused on the accuracy of the answers \cite{nguyen2019overcoming,liu2023parameter,zhan2020medical}
, where most MedVQA responses consist of a simplistic answer lacking detailed explanations or rationale, unlike in real-world scenarios where doctors not only provide answers but also explain their reasoning, professional considerations, and potential contradictions to derive a more comprehensive diagnostic insight. 
Besides, real-world diagnostics often rely on the combined experience of multiple doctors, as a single doctor's diagnosis may be biased by personal experience and may not be sufficiently accurate.
In the multimodal Chain of Thought (CoT), answering VQA questions involves providing an answer as well as a corresponding reasoning path (rationale). 
The generation of this rationale helps to improve the accuracy of the language model.
Inspired by real-world practices and multimodal CoT \citet{zhang2023multimodal,zheng2023ddcot}, integrating this paradigm into Med-VQA can enhance both the accuracy and interpretability of responses. However, implementing it faces several challenges: (1) Previous CoT methods required manual annotation of fundamental rationales, which is time-consuming, costly, and challenging to ensure consistency and completeness \cite{zhang2023multimodal,zheng2023ddcot}. (2) Reliance on a single expert model can lead to misleading conclusions. (3) Multimodal CoT has limited depth in understanding the intents of images and texts, which can restrict its effectiveness in medical contexts \cite{zhang2023multimodal}.

To address the aforementioned issues, we introduce MedCoT, a hierarchical expert-verified model for Med-VQA. Firstly, the Initial Specialist proposes preliminary diagnostic rationale based on the medical visual and text query. The Follow-up Specialist then reviews these rationales, categorizing them as valid or invalid; valid rationales are retained, while invalid ones are reassessed. 
Finally, the locally implemented Diagnostic Specialist, consisting of a sparse Mixture of Experts (MoE) model functioning as a multimodal language model, casts votes to deliver the definitive diagnosis.
Leveraging a hierarchy of expertise, MedCoT consistently outperforms state-of-the-art (SoTA) Med-VQA methods across four extensive datasets, demonstrating impressive generalizability and interpretability, as shown in \autoref{fig1}.
Our study makes three significant contributions:

\begin{itemize}
    \item We have conducted an in-depth analysis of the challenges and insights associated with generating rationales in multimodal CoT. Our findings highlight that single specialist often fails to provide clear verifications and are more prone to errors when addressing questions about specific organs.
    \item Inspired by real-world diagnostics, we developed the hierarchical expert-verified MedCoT, which does not require manually annotated rationales. This involves three tiers of expert verification: initial, follow-up, and diagnosis. MedCoT not only provides more accurate answers but also offers refined rationales.
    \item In the diagnosis stage, we designed a sparse MoE that includes majority voting. This framework's multiple specialized experts efficiently and accurately interpret the intents of medical images and texts, enabling the Diagnostic Specialist to provide precise responses.
    
    
\end{itemize}

\vspace{1em}



\begin{figure*}
\centering
\includegraphics[width=\textwidth]{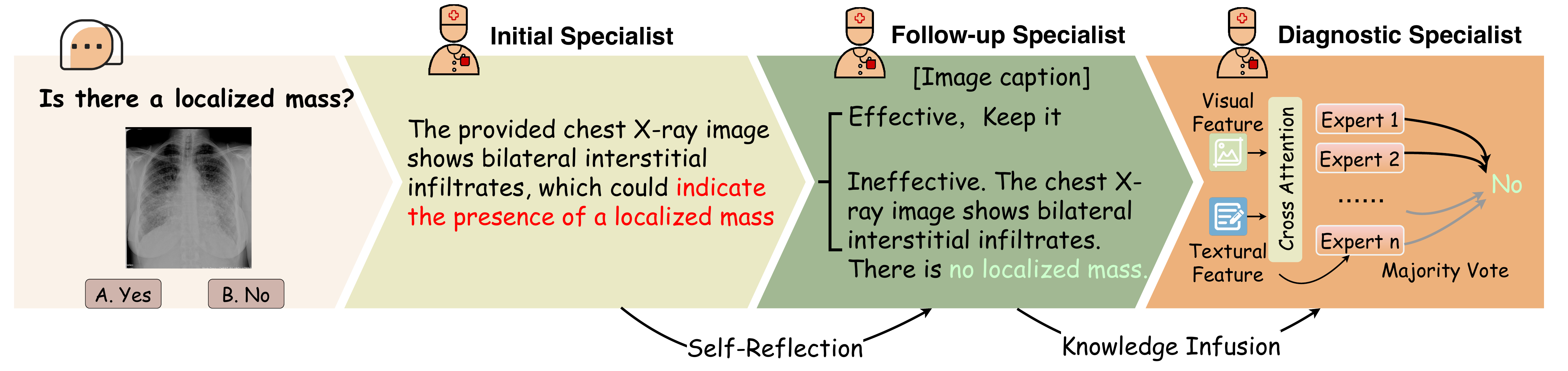}
\caption{
The MedCoT pipeline begins with an Initial Specialist receiving a medical question and image to generate a preliminary rationale. This rationale may have flaws (indicated in red), which are then reviewed by the Follow-up Specialist. If the rationale is deemed effective, it is retained; otherwise, it is reconsidered and a new rationale (indicated in green) is generated, along with an image caption. These elements are then integrated into the Diagnostic Specialist. Informed by all contexts, the Diagnostic Specialist, a multimodal language model with a designed sparse MoE structure, delivers the final diagnostic outcome (answer).
} 
\label{MedCoT pipeline}
\vspace{-1em}
\end{figure*}

\section{Related Work}
\vspace{-0.5em}
\subsection{Med-VQA}
VQA is a multimodal task in computer vision and natural language processing, aimed at responding to queries about images in natural language \cite{ben2019vqa,he2020pathvqa,ren2020cgmvqa}. It involves feature extraction, fusion, and inference to comprehend multimodal intents and manage feature processing. Med-VQA extends VQA into the medical domain, where robust medical knowledge is crucial for answering domain-specific questions \cite{liu2023parameter}, thus complicating feature extraction. Innovations such as Nguyen et al.'s MEVF leverage unsupervised CDAE and meta-learning to initialize weights specifically for Med-VQA \cite{nguyen2019overcoming}. Zhan et al. built upon this by developing a conditional reasoning framework to handle different types of questions \cite{zhan2020medical}, while Eslami et al. successfully implemented the CLIP model as a visual encoder, proving its effectiveness in this context \cite{eslami-etal-2023-pubmedclip}. LLaVA-Med utilizes GPT-4 and a novel curriculum learning approach for training on biomedical images \citet{li2024llava}, significantly enhancing Med-VQA capabilities. While capable of interactive dialogue, its responses do not focus on the reasoning paths leading to the answers. MedCoT differs from the aforementioned methods by not only providing precise answers but also offering reasoning paths (rationale). Moreover, its validity is confirmed through Hierarchical Expert verification, aligning more closely with real-world medical scenarios.

\subsection{Multimodal CoT}
CoT reasoning with Large Language Models (LLMs) has shown success in natural language processing. Multimodal CoT combines visual information with traditional textual CoT, integrating comprehensive data to perform reasoning tasks \cite{zhang2023multimodal,zheng2023ddcot}.
Groundbreaking works in multimodal CoT  \cite{zheng2023ddcot,zhang2023multimodal,lu2022learn,lu2023chameleon,zhang2023llama} are first examined on the ScienceQA dataset. ScienceQA includes multimodal scientific questions along with annotated rationales \cite{lu2022learn}.
MM-CoT developed a two-stage framework based on ScienceQA that trains models to generate rationales from annotations, which are then used to form final answers \cite{lu2022learn}. 
With the increasing integration of open-world knowledge in LLMs, research is focusing on equipping these models with visual modalities to tackle complex visual and multimodal challenges. For instance, DDCoT \cite{zheng2023ddcot}, introduces role-specific Chains of Thought that decompose questions into subproblems and use LLMs to recombine principles, enhancing accuracy and addressing language illusions in multimodal contexts.
Inspired by these advancements, we aim to adapt multimodal CoT reasoning to the medical field, aiming to improve the explainability and accuracy of Med-VQA.

\subsection{MoE}
MoE optimizes learning and prediction by combining multiple expert networks and using a gating network to determine which experts are activated based on the given input \cite{zhang2024scalable,fedus2022switch}. Sparse MoE, a variant of the MoE model, activates only a few experts during each prediction, thus efficiently utilizing computational resources and enhancing scalability \cite{shazeer2016outrageously}.
Sparse MoE models have been independently explored within the context of conditional computation in both computer vision and natural language processing domains \citet{jacobs1991adaptive,fedus2022review}. Conditional computation aims to increase the number of model parameters without proportionally increasing computational costs. This is achieved by selectively activating only the relevant parts of the model based on input-specific factors \cite{shazeer2016outrageously}. Sparse MoE models employ a learned gating mechanism that activates only a subset of experts, specifically \( k \) out of \( N \) experts, for a given input. This allows for the selection of either all experts or just a sparse mix, optimizing resource usage \citet{lepikhin2020gshard}. 


\begin{figure}[t]
\centering
\includegraphics[width=0.5\textwidth]{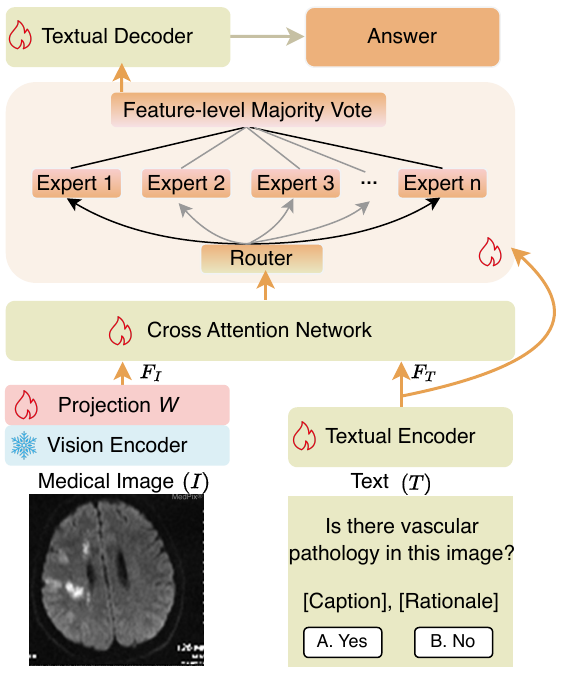}
\caption{Diagnostic Specialist Pipeline. After passing through a visual encoder, medical images yield visual features. Contextual textual information—including captions, rationales, and options—is processed by a text encoder to obtain textual features. These are then subjected to cross-attention for feature integration, producing combined features. These integrated features, along with textual features, are input into a Sparse MoE structure. Here, multiple specialized experts thoroughly understand the intents of both the image and text. The insights are then fed into a textual decoder, which decodes the information to produce the final answer.
} 
\label{Diagnostic}
\vspace{-1em}
\end{figure}
\section{Methodology}

\subsection{Preliminaries}
Throughout this paper, we model the Med-VQA task within a multimodal CoT framework as follows: The framework takes an image \( I \) and a question \( Q \) as inputs, and outputs a reasoning rationale \( R \). This rationale \( R \) is subsequently used to generate an answer \( A \). This paradigm ensures that the process is transparent, providing a traceable path from input to conclusion, which is essential for both validating the results and improving user trust in the framework's diagnostic capabilities. We can model the Med-VQA task within a multimodal CoT as follows:
\begin{equation}
    \min_{f, g} \mathbb{E}_{(I, Q, A^*) \sim \text{Data}} \left[ L \left( g \left( f(I, Q), I, Q \right), A^* \right) \right].
\end{equation}

\( f \) is responsible for generating a rational and helpful reasoning rationale \( R \) (Initial and Follow-up Specialists), while \( g \) uses this rationale to generate the final answer \( A \) (Diagnostic Specialist). The rationale \( R \) is derived from the Initial Specialist assessments and self-reflection by the Follow-up Specialist. The final answer \( A \) is determined by a Diagnostic Specialist through a loss function \( L \), which measures the discrepancy between the predicted answer \( A \) and the true answer \( A^* \).


\subsection{Initial Specialist}
In the initial diagnosis phase, we cue the LLMs to act as the primary rationale Diagnostic Specialist. 
We prompt the LLMs with the instruction: 
\textit{"Please proceed with a step-by-step analysis and provide a rationale"} ($prompt_{\hat{i}}$).
This is done to guide the LLMs in performing a detailed, step-by-step reasoning process. 
The textual rationale obtained from this is represented as \( R_{\hat{i}} = LLMs(T, I, prompt_{\hat{i}}) \), where \( T \) and \( I \) denote the text and image inputs, respectively. \( T \) includes textual context such as the question \( Q \) and options.
\( prompt_{\hat{i}} \) is the specific prompting strategy used to elicit the rationale. For further technical details about the prompt, please refer to the appendix.

For instance, as shown in \autoref{MedCoT pipeline}, for the question "Is there a localized mass?", we obtain a highly interpretable rationale (for the final diagnostic outcome): "The provided chest X-ray image shows bilateral interstitial infiltrates, which could indicate the presence of a localized mass".
\subsection{Follow-up Specialist}
In the follow-up diagnosis phase, we instruct LLMs to conduct self-reflection reasoning and test within the problem's context to identify effective rationales, retain them, and reconstruct ineffective ones to generate accurate rationales. Specifically, we prompt the LLMs with:
\textit{"Please judge whether this rationale is effectively valid for the question and image. If it is effective..., If the existing rationale is Ineffective..."} ($prompt_{\hat{f}}$). For the complete prompt, please refer to the appendix. We can define the Self-Reflection reasoning of the Follow-up Specialist using the following formula:
\begin{equation}
\begin{small}
R_{\hat{f}} = 
\begin{cases} 
R_{\hat{i}} & \text{if } R_{\hat{i}} = \text{Effective} \\
\text{LLMs}(T, I, {prompt}_{\hat{f}}) & \text{if } R_{\hat{i}} = \text{Ineffective},
\end{cases}
\end{small}
\end{equation}
where $R_{\hat{f}}$ is Follow-up Specialist rationale. This process helps us obtain the textual rationale needed for the diagnostic analysis, as shown in \autoref{MedCoT pipeline}.

To infuse the Diagnostic Specialist with more knowledge and bridge the gap between image and text, we utilize the Follow-up Specialist to generate image captions. This process helps to reduce the modality gap, effectively channeling this knowledge into the Diagnostic Specialist. For detailed caption prompts, please refer to the appendix.


\subsection{Diagnostic Specialist}
We employ the designed model based on multimodal T5 combined with sparse MoE to serve as the Diagnostic Specialist, as shown in \autoref{Diagnostic}. The Diagnostic Specialist receives enriched textual context and medical imaging information to generate the final diagnostic outcome.
\subsubsection{Multimodal T5}
\autoref{Diagnostic} shows the structure of multimodal T5, including the \textit{TextualEncoder}, \textit{VisualEncoder}, \textit{Cross-Attention Network}, sparse MoE, and the \textit{TextualDecoder}. Here are the network details:

\textit{TextualEncoder} transforms natural language input \( {T} \) into the textual feature space \( F_T \in \mathbb{R}^{n \times d} \), and \textit{VisualEncoder} converts the input image \( I \) into visual features \( F_I \in \mathbb{R}^{m \times d} \). 
Here, \( n \) signifies the length of the input language text, \( d \) the dimensionality of hidden features, and \( m \) the count of image patches.
Upon obtaining the textual representation \( F_T \) and visual representation \( F_I \), our model leverages the \textit{Cross-Attention Network} for modality interaction. This network computes the attention-guided visual feature \( H_{V}^{\text{att}} \in \mathbb{R}^{n \times d} \), which selectively captures relevant visual features in response to the textual query, as delineated in the operation:
\begin{align}
        H_{V}^{\text{att}} &= \text{Softmax}\left(\frac{QK^{\top}}{\sqrt{d}}\right)V,
\end{align}
where $Q$, $K$, $V$ correspond to the query, key, and value, derived from $F_T$, $F_I$, $F_I$, respectively.

Once the attention-guided visual feature \( H_{V}^{\text{att}} \) and the textual representation \( F_T \) are obtained, we construct the MoE to dynamically amalgamate them, resulting in \( F_F = \text{MoE} (H_{V}^{\text{att}}, F_T) \). Details of the MoE are provided in the following section.
$F_{\text{ F}}$ is input into the \textit{TextualDecoder} to generate answer
$A = \text{TextualDecoder}(F_{\text{F}})$, as shown in \autoref{Diagnostic}. 


In the training, refinements enable predicted answers (A) to more accurately approximate label answers. Specifically,The model $f$ with input maximizes the likelihood of the correct sequence $Y = {A}$. The loss function $L$, which is the negative log-likelihood over all tokens, is given by: $L = -\sum_{n=1}^{N} \log p(Y_n | X, Y_1^{n-1})$, where $N$ is the number of tokens, and $p(Y_n | X, Y_1^{n-1})$ is the probability of predicting the correct $n$-th token in $Y$. 
\subsubsection{MoE}

In the multimodal CoT, a crucial step is understanding the intent of both the image and the text and responding accordingly. 
However, previous methods primarily utilized gates for integration, where the gate function \( \lambda = \text{Sigmoid}(W_l F_T + W_v H_V^{\text{att}}) \) weights the importance of the image relative to the source text, with \( W_l \) and \( W_v \) as learnable parameters (see Appendix) \cite{zhang2023multimodal, zheng2023ddcot}.
Which, according to our experiments, shows that the gate is insufficient (\autoref{ablation}).
Therefore, MedCoT proposes constructing a MoE for the integration process.

The Sparse MoE implements a top-k sparse mixture of experts \cite{fedus2022switch}, leveraging multiple Sparse Experts to specialize in processing complex Med-VQA data. 
This module dynamically selects the top-k experts for each input based on gating scores, as shown in \autoref{Diagnostic}. 

After obtaining the outputs from the experts, we use Feature-level Majority Vote to aggregate their outputs.
The weight of each expert is calculated using the following formula:
\begin{align}
    \text{W}_i = \text{softmax}(\text{V}^{\text{top k}})_i = \frac{e^{\text{V}^{\text{top k}}_i}}{\sum_{j=1}^{k} e^{\text{V}^{\text{top k}}_j}},
\end{align}


where \(\text{W}_i\) is the weight of the \(i\)-th selected expert, and \(\text{V}^{\text{top k}}_i\) is the score of the \(i\)-th selected expert.
For each feature \( F_f \), the final result of Feature-level Majority Vote is calculated by weighted averaging the outputs of all selected experts:
\begin{align}
    {E}_{F_f} = \sum_{i=1}^{k} \text{W}_i \cdot E_{i, F_f} ,
\end{align}
where \({E}_{F_f}\) is the value of the final result for feature \( F_f \), and \(E_{i, F_f}\) is the output of the \(i\)-th selected expert for feature \( F_f \). Then,
\(\lambda = \text{Sigmoid}({E}_{F_f}) \).
Finally, this results in \( F_f \) are as follows:
\begin{align}
    F_{\text{F}} &= (1-\lambda) \cdot F_{\text{T}} + \lambda \cdot H_{V}^{\text{att}}.
\end{align}

The sparse MoE network allows each selected expert to handle data they specialize in, as demonstrated in \autoref{rad}, which shows experts proficient in addressing head-related issues.







\begin{figure}[t]
\centering
\includegraphics[width=0.4\textwidth]{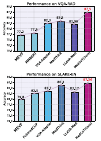}
\caption{
MedCoT is compared with various SoTA methods on closed questions on the VQA-RAD and SLAKE-EN datasets. MedCoT not only achieves SoTA accuracy in answers but also provides reasoning paths (rationale). The metric used is Accuracy (\%).
} 
\label{performance}
\vspace{-1em}
\end{figure}

\begin{figure*}
\centering
\includegraphics[width=\textwidth]{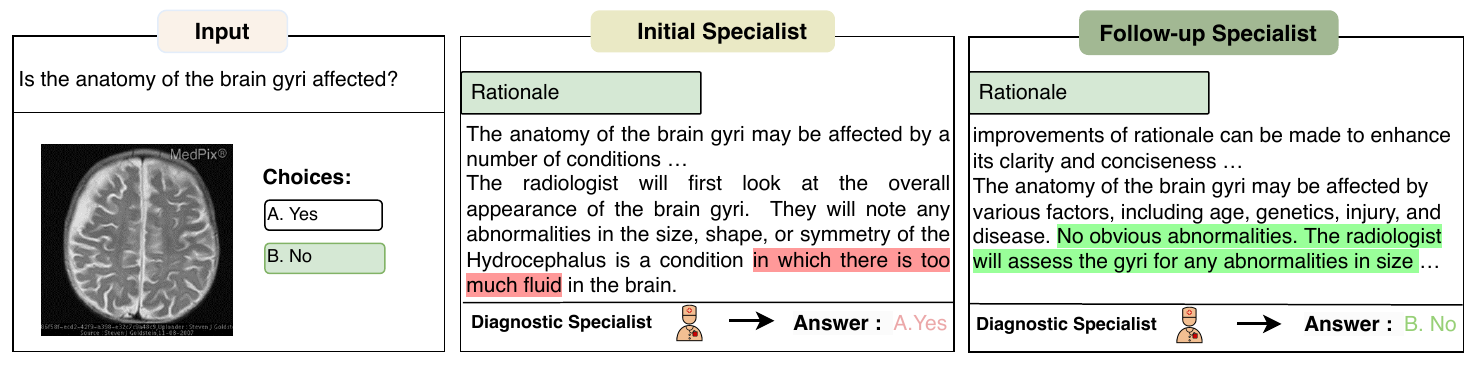}
\caption{
The MedCoT pipeline begins with an Initial Specialist receiving a medical question and image to generate a preliminary rationale. This rationale may have flaws (indicated in red), which are then reviewed by the Follow-up Specialist. If the rationale is deemed effective, it is retained; otherwise, it is reconsidered and a new rationale (indicated in green) is generated, along with an image caption. These elements are then integrated into the Diagnostic Specialist. Informed by all context, the Diagnostic Specialist, a multimodal language model with a designed sparse MoE structure, delivers the final diagnostic outcome (answer).
} 
\label{case1}
\end{figure*}

\section{Experiments}

\subsection{Experimental Setting}
In MedCoT framework, the encoder and decoder from Flan-T5 \cite{khashabi2020unifiedqa,raffel2020exploring} are integrated as TextualEncoder($\cdot$) and TextualDecoder($\cdot$), respectively. Additionally, DETR \cite{carion2020end} is employed as VisualEncoder($\cdot$). 
Our Diagnostic Specialist model was trained 100 epochs with a learning rate of $8e-5$ and a batch size of 8. 
To demonstrate the effects of MedCoT, four benchmark datasets are used for validation in the medical VQA domain: VQA-RAD \cite{lau2018dataset}, SLAKE-EN \cite{liu2021slake}, Med-VQA-2019 \cite{abacha2019vqa}, and PathVQA \cite{he2020pathvqa}, with detailed statistics provided in Appendix. 
All experiments were conducted using PyTorch \cite{paszke2019pytorch} and HuggingFace \cite{wolf2020transformers}, implemented on 4 NVIDIA GEFORCE RTX 3090 GPUs. Accuracy is utilized as the evaluation metric. 
For LLMs, Gemini Pro 1.5 version is used for our Initial Specialist and Follow-up Specialist. The more experimental details can be found in the Appendix.

\begin{figure}
\centering
\includegraphics[width=0.5\textwidth]{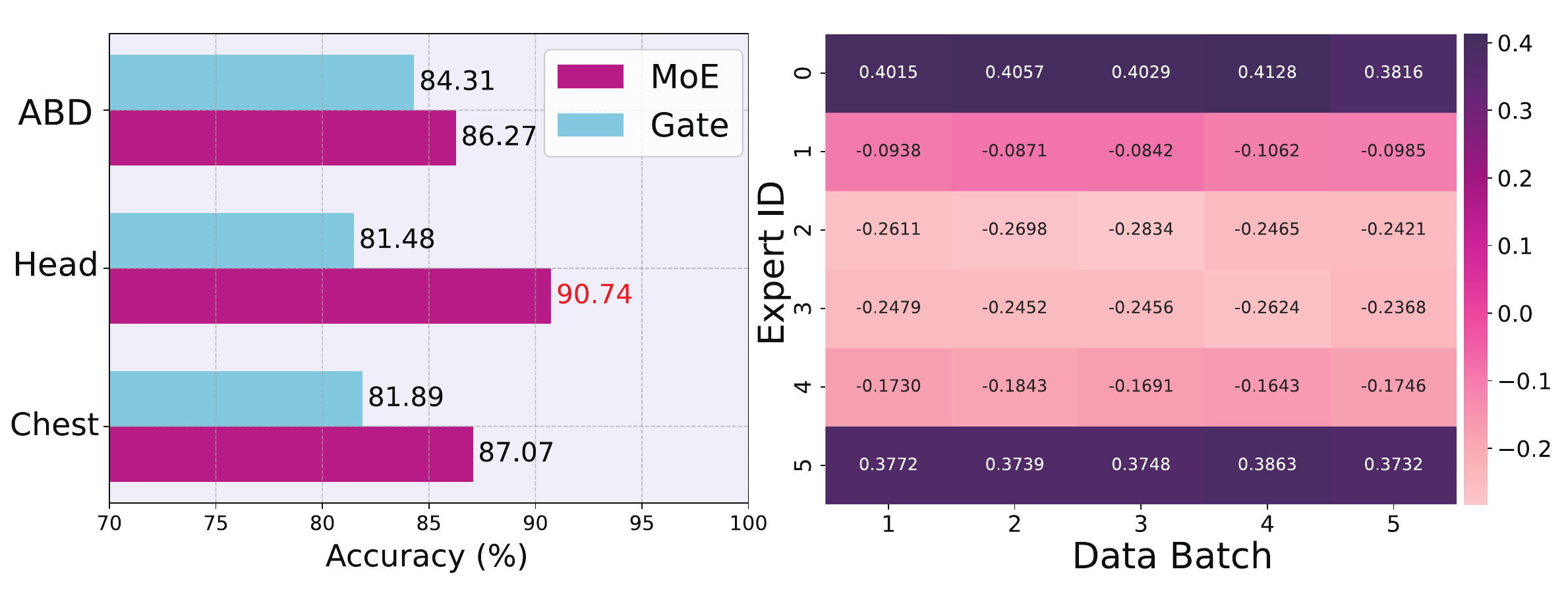}
\caption{
The Diagnostic Specialist's sparse MoE shows varying accuracy levels for different organ-related questions in VQA-RAD. 
'ABD' represents abdominal-related questions, 'Head' refers to head-related questions, and 'Chest' refers to chest-related questions.
It can be observed that head-related questions saw an improvement of nearly 10 \%. We visualized the weights of the experts (right figure). Notably, in the top 2 expert selections, the model chose Expert 0 and Expert 5 to understand the intents of the "head" image and text. 
} 
\label{rad}
\end{figure}

\begin{figure*}
\centering
\includegraphics[width=0.9\textwidth]{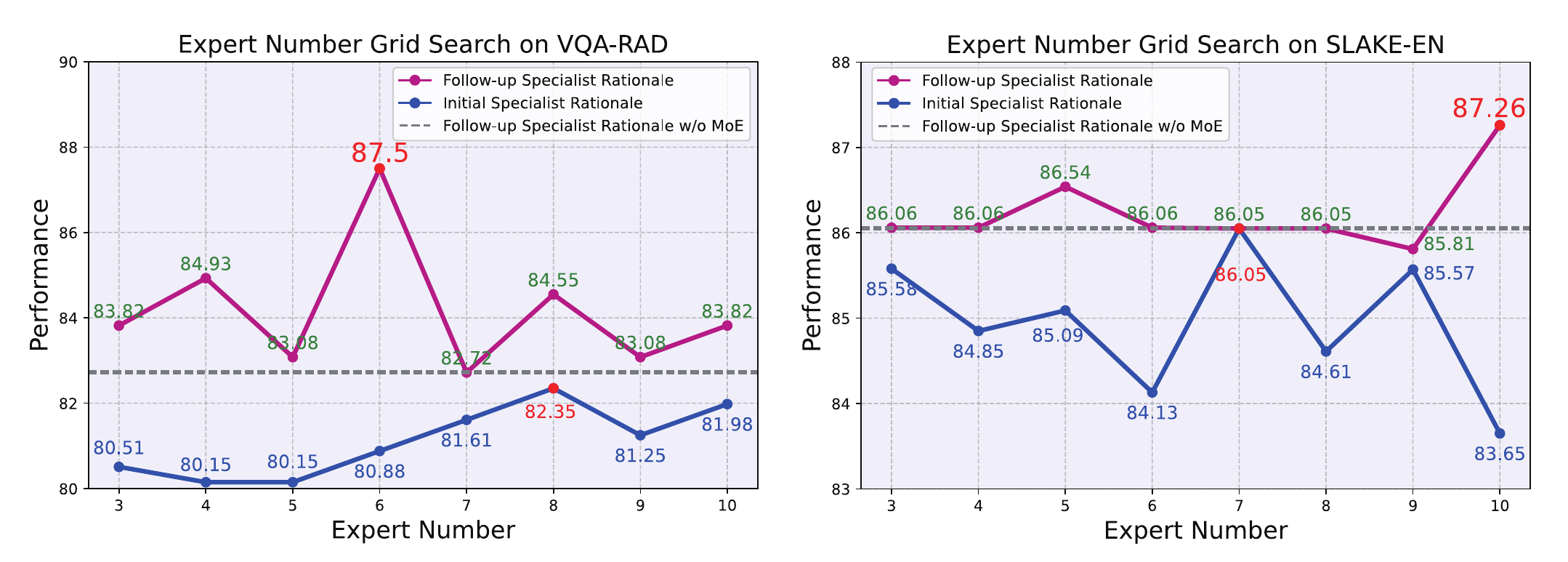}
\caption{The expert number grid search on two datasets. The blue line represents the results from training with Initial Specialist rationales and grid search of expert numbers in the Diagnostic Specialist. The purple line represents results from using the Follow-up Specialist rationales and grid searching expert numbers. 
The gray line represents the results of the Diagnostic Specialist using Follow-up Specialist rationales, conducted without the sparse MoE.
} 
\label{zhexiantu-all}
\vspace{-1em}
\end{figure*}

\subsection{Main Results}
We evaluate the performance of MedCoT on the VQA-RAD and SLAKE-EN datasets, benchmarking them against established models like MEVF \cite{nguyen2019overcoming}, MMBERT \cite{tiong-etal-2022-plug}, PubMedCLIP \cite{eslami-etal-2023-pubmedclip}, VQA-Adapter \cite{liu2023parameter}, MedThink \cite{gai2024medthink}, LLaVA-Med \cite{li2024llava}. 

Our performance evaluation is divided into two parts, focusing separately on closed-end and open-end questions. Closed-end questions, structured as multiple-choice questions with a single correct answer, are assessed using accuracy as the performance metric, as shown in \autoref{performance}.
In facing closed-end questions, 
MedCoT surpasses a range of SoTA methods on the VQA-RAD and SLAKE-EN datasets. Notably, MedCoT achieved improvements of 27.21\% and 14.66\% over Gemini on the two datasets, demonstrating the unreliability of a single model. Besides, MedCoT, with a fine-tuning size of approximately 256M parameters, outperforms the 7B parameter LLaVA-Med (trained on extensive medical data), exceeding it by 5.52\% and 4.09\% on two datasets, respectively. Moreover, compared to previous methods, MedCoT clearly displays the reasoning paths (rationale), as illustrated in \autoref{MedCoT pipeline}.
More comparative method results can be seen in  Appendix.

In contrast, open-end questions allow for a range of answers due to their inherent nature. The answers generated by MedCoT are difficult to match precisely against the dataset. Therefore, we employ text generation metrics such as Rouge and BLEU to evaluate MedCoT's performance.
We conducted experiments on the open-end VQA-RAD and SLAKE-EN, with results shown in the Appendix. 
MedCoT demonstrated higher Rouge and BLEU scores on the VQA-RAD and SLAKE-EN dataset, surpassing MedThink \cite{gai2024medthink}.
Besides, MedCoT also showed higher scores on the SLAKE-EN.

Additionally, we evaluated MedCoT's performance on the Med-VQA-2019 and PathVQA datasets, as shown in Appendix. The results indicate that MedCoT consistently achieves SoTA results compared to the majority of SoTA methods.

\begin{table}
\centering
\small
\setlength{\extrarowheight}{0pt}
\addtolength{\extrarowheight}{\aboverulesep}
\addtolength{\extrarowheight}{\belowrulesep}
\setlength{\aboverulesep}{0pt}
\setlength{\belowrulesep}{0pt}
\caption{Ablation Study on MedCoT}
\label{xiaorong}
\begin{tabular}{cc|cc} 
\toprule
Follow-up & MoE & VQA-RAD                                        & SLAKE-EN                                        \\ 
\hline
          &     & 77.57                                          & 83.17                                           \\
   \greencheck       &     & 82.72                                          & 86.05                                           \\
          &  \greencheck   & 80.88                                          & 83.65                                           \\
  \greencheck        &  \greencheck   & {\cellcolor[rgb]{1,0.875,0.757}}\textbf{87.50} & {\cellcolor[rgb]{1,0.875,0.757}}\textbf{87.26}  \\
\bottomrule
\end{tabular}
\end{table}

\subsection{Ablation Study}
\label{ablation}
\noindent\textbf{Effects of Follow-up Specialist}
To validate the effectiveness of the Follow-up Specialist, we compared the results of experiments involving only the initial and diagnostic specialists with those from the complete MedCoT. As shown in \autoref{xiaorong}, across two medical datasets, there is a significant performance loss when the Follow-up Specialist is removed. For instance, on the VQA-RAD dataset, performance dropped from 87.50\% to 80.88\%, a decrease of 6.62\%. This demonstrates the effectiveness of the Follow-up Specialist. 

Besides, we conducted experiments involving only the initial and diagnostic specialists, 
bypassing the self-reflection of the Follow-up Specialist. 
In all cases involving varying numbers of experts, the results without the self-reflection were consistently lower than those with rationales refined by the Follow-up Specialist’s reflection,  and even lower than those from a Diagnostic Specialist that had undergone self-reflection but was lacking the MoE component, as shown in \autoref{zhexiantu-all}. 
This underscores the importance of the self-reflection provided by the Follow-up Specialist.
Additionally, we conducted zero-shot experiments using both the initial and Follow-up Specialist. As shown in the appendix, these results further confirm the effectiveness of the Follow-up Specialist.

\noindent\textbf{Effects of MoE}
To validate the effectiveness of the MoE, we compared the performance with and without the MoE. As shown in \autoref{xiaorong}, there is a significant performance drop across all datasets without MoE. For instance, in the VQA-RAD, the performance decreased from 87.50\% to 82.72\%, a loss of 4.78\%. This indicates that MoE plays a crucial role in Diagnostic Specialist.
As can also be seen from \autoref{zhexiantu-all}, lacking MoE, in most expert number scenarios, the performance is weaker compared to MedCoT equipped with Sparse MoE.

Additionally, we conducted experiments for each organ-related question category within the VQA-RAD and SLAKE-EN, as shown in \autoref{rad}. 
It is evident that in the majority of organ-related questions, methods employing MoE outperform those using the gating mechanism. 
Notably, the Gate mechanism, resembling as a single-expert system, tends to falter with head-related questions, where it performs the worst. For such questions in the VQA-RAD, methods using MoE exceeded those with gates by 10\%, further emphasizing MoE's effectiveness.
We visualized the weights of MoE, as shown in the \autoref{rad} (right figure), revealing that Experts 0 and 5 primarily handle head-related issues. This demonstrates that these two experts dynamically process and understand the intents of medical images and texts more effectively than the gating. 
Similar results can also be observed in the experiments conducted on the SLAKE-EN, as shown in Appendix.


\noindent\textbf{Grid Search}
We conducted a parameter search experiment for the hyperparameters in the sparse MoE, such as the number of experts and the \( k \) value. The results are shown in Appendix. The experiment revealed that the optimal number of experts varies for different datasets. Specifically, the best number of experts for VQA-RAD, SLAKE-EN, Med-2019 and PathVQA are 6, 10, 5, and 5, respectively. Regarding the \( k \) value, the optimal value for all datasets was consistently 2, as illustrated in Appendix.
\subsection{Discussion}

\autoref{MedCoT pipeline} and \autoref{case1} illustrate cases where the Initial Specialist provides a rationale, the Follow-up Specialist makes corrections, and the Diagnostic Specialist delivers the final, accurate diagnosis. For instance, in \autoref{case1}, the Initial Specialist, influenced by the illusions of the LLMs, mistakenly observes non-existent brain fluid and diagnoses the brain as being affected by gyri. However, after the self-reflection by the Follow-up Specialist, it is clarified that no clear fluid was observed. Ultimately, the Diagnostic Specialist, using the rationale from the Follow-up Specialist and considering the full context, arrives at the correct diagnosis.

Appendix provides an example where the limitations of LLMs affect the ability to accurately diagnose certain cases. 
The question posed is whether there is pneumomediastinum. The Initial Specialist, based on observations, affirms its presence, and the Follow-up Specialist concurs, leading to a unanimous agreement. However, due to the limitations of the LLMs, these rationales are incorrect, ultimately leading to an erroneous answer.

\section{Conclusion}
In this paper, we propose an effective hierarchical expert reasoning chain method for Med-VQA, named MedCoT. This method is based on two insights: 1) Med-VQA should have a clear reasoning path; 2) Med-VQA scenarios should be reviewed by multiple experts to arrive at a conclusion. Specifically, the process involves initial experts providing preliminary diagnostic rationales based on medical visual questions. Follow-up experts then review these rationales for validity, retaining the effective ones and reassessing the ineffective ones. Finally, a locally deployed Diagnostic Specialist, consisting of a sparse MoE that conducts a vote, then provides the definitive diagnosis. Experimental results on multiple Med-VQA datasets show that MedCoT outperforms existing SoTA techniques, significantly surpasses recent methods, and demonstrates excellent interpretability for final diagnosis.

\newpage
\section*{Limitation}
A limitation is that the performance of MedCoT is influenced by the hallucinations of the LLMs used by the Initial and Follow-up Specialist. 
Although self-reflection and Hierarchical Expert design can mitigate some issues with LLMs' hallucinations, it must be acknowledged that the problem is not completely resolved. As shown in Appendix, MedCoT is still susceptible to hallucination risks. Researching methods to suppress hallucinations is a potential topic for further study.
In this work, the Gemini-Pro model was employed. If Med-Gemini becomes available, MedCoT could be further enhanced.
Moreover, MedCoT could inspire future paradigms that integrate proprietary commercial LLMs with local models. 
By utilizing desensitized information to prompt the extensive knowledge and reasoning capabilities of LLMs, the generated rationales could be combined with local models for further diagnostic analysis, enhancing both interpretability and accuracy. 

Another limitation is that compared to single-model methods, MedCoT may be more time-consuming. However, the hierarchical expert approach aligns more closely with real-world medical diagnostics and provides clear diagnostic pathways as well as more accurate answers, making the additional time worthwhile.
\section*{Acknowledgements}
This work is supported by the National Natural Science Foundation of China (Grant No. 62106222), the Natural Science Foundation of Zhejiang Province, China (Grant No. LZ23F020008), and the Zhejiang University-Angelalign Inc. R\&D Center for Intelligent Healthcare.
This work is also supported by Jiawei Du’s A*STAR Career Development Fund (CDF) C233312004.
\bibliography{main}
\newpage

\end{document}